\documentclass[10pt, a4paper]{article}
\usepackage{microtype}
\usepackage{dirtytalk}
\usepackage{amsmath}
\usepackage{booktabs}
\usepackage{adjustbox}
\usepackage{array}
\usepackage{tabulary}
\usepackage{todonotes}
\setuptodonotes{inline}
\usepackage{enumerate}
\usepackage{enumitem}
\usepackage{tcolorbox}
\usepackage{tabularx}
\usepackage{subcaption}
\usepackage{covington}
\usepackage[final]{lrec2026} 
\definecolor{cbBlue}{RGB}{31, 120, 180}
\definecolor{cbGreen}{RGB}{51, 160, 44}
\definecolor{cbOrange}{RGB}{255, 127, 0}
\newcommand{\hla}[1]{\textcolor{cbBlue}{#1}}
\newcommand{\hlb}[1]{\textcolor{cbOrange}{#1}}

\tcbset{
    colback=gray!5,
    colframe=black!50,
    arc=2mm,
    boxrule=0.5pt,
    left=2mm,
    right=2mm,
    top=1mm,
    bottom=1mm,
    fontupper=\scriptsize, 
    fonttitle=\scriptsize\bfseries, 
}

\DeclareMathOperator{\MeanLP}{MeanLP}

\setexampleoptions{  
  fs={\normalfont\itshape},
  fspreamble={\normalfont\upshape},
  subnumberformat={a.},
  addsubnumbersep={-4pt},
}

\title{A Dataset for Probing Translationese Preferences in English-to-Swedish Translation}

\name{Jenny Kunz \quad Anja Jarochenko \quad Marcel Bollmann} 

\address{Department of Computer and Information Science \\
Linköping University \\
         \texttt{jenny.kunz@liu.se} \quad \texttt{anjja777@student.liu.se} \quad \texttt{marcel.bollmann@liu.se}\\}

\abstract{
Translations often carry traces of the source language, a phenomenon known as \textit{translationese}. We introduce the first freely available English-to-Swedish dataset contrasting translationese sentences with idiomatic alternatives, designed to probe intrinsic preferences of language models. It includes error tags and descriptions of the problems in the original translations. 
In experiments evaluating smaller Swedish and multilingual LLMs with our dataset, we find that they often favor the translationese phrasing. Human alternatives are chosen more often when the English source sentence is omitted, indicating that exposure to the source biases models toward literal translations, although even without context models often prefer the translationese variant. Our dataset and findings provide a resource and benchmark for developing models that produce more natural, idiomatic output in non-English languages. 
 \\ \newline \Keywords{Translationese, Idiomaticity, Machine Translation, Evaluation, Minimal Pair Probes} }

\begin{document}

\maketitleabstract

\section{Introduction}
\label{sec:introduction}

Translations have long been known to carry traces of the source language and differ in style and features from texts originally written in the target language; a phenomenon commonly called \textit{translationese}~\citep{gellerstam86}. 
Research shows that translationese is particularly strong in machine translation, leading to simplified and less varied language marked by reduced lexical and morphological richness~\citep{vanmassenhove-etal-2021-machine}. Such output is often easily distinguishable from original text~\citep{koppel-ordan-2011-translationese, li-etal-2025-lost}.
Recent work suggests that LLMs produce less literal translations~\citep{raunak-etal-2023-gpts}. However, even though their outputs show increased lexical diversity compared to specialized machine translation systems, they can still be reliably distinguished from human-written text~\citep{kong-macken-2025-decoding}. 

For many languages, it is common to use translated datasets especially for LLM~evaluation \citep{nielsen-2023-scandeval, bandarkar-etal-2024-belebele} and instruction tuning \citep{li2023bactrianxmultilingualreplicableinstructionfollowing, dac2023okapi, holmstrom-doostmohammadi-2023-making}, as there is often no practical alternative. In addition, for low-resource languages, but to a lesser extent even for high-resource languages such as English, large portions of modern web-crawled corpora that constitute the training data of LLMs are translations \citep{thompson-etal-2024-shocking}.  
It is therefore important to analyze this problem and assess its extent, in order to move towards models that produce natural, idiomatic output in non-English languages.

\citet{anonymous2025} 
investigate conventionalized idiom knowledge and translationese in LLM outputs for Swedish. Analyzing sentence pairs from a book on translationese~\citep{katourgi2022svenskan}, which contains translations with translationese phrasing alongside idiomatic alternatives suggested by the author, they examine whether the model assigns higher probability to the translationese version or to the idiomatic alternative. Building on this work, we construct a dataset for probing translationese preferences. Our dataset is similar in its setup, but addresses several limitations of the previous dataset: it is fully open under a permissive license, includes the English source sentence with preceding context to also study preferences in a translation context, and provides annotations tagging the type of translationese, enabling more fine-grained analysis of model behavior. To our knowledge, this is the first freely available dataset explicitly contrasting translationese with idiomatic alternatives for Swedish.

We provide a detailed description of the dataset, including both qualitative and quantitative analyses of translationese examples in English-to-Swedish translations produced by a smaller specialized machine translation system. We also compare these to translations generated by a state-of-the-art LLM, highlighting that while it does not fully resolve any specific problem, it is much better at generating more idiomatic
words and phrasings. 

In experiments probing the intrinsic preferences of smaller multilingual LLMs with our dataset, findings show a strong bias toward the translationese phrasing. Notably, models favor the human alternative more often when omitting translation context in the prompt, indicating that exposure to the English source sentence biases them toward the more literal, translationese wording. However, increasing the context window helps the model become less biased towards the translationese variant.

\paragraph{Release} We release our dataset on HuggingFace under \url{https://huggingface.co/datasets/liu-nlp/translationese-opensubtitles}. A full version including all annotations with further explanations, including the code to reproduce our experiments, is available on GitHub: \url{https://github.com/jekunz/translationese}. 

\section{Background}
\label{sec:background}

\paragraph{Translationese} Translated texts often preserve features of the source language while avoiding target-language constructions that diverge more strongly from the source language. This is just one aspect of a phenomenon known in both human and machine translation as \textit{translationese}~\citep{gellerstam86}. Translationese is not generally a sign of poor translation quality 
\citep{Gellerstam2005} but the resulting texts have properties that are different from idiomatic language. As \citet{baker} shows, translated texts tend to be more explicit than their originals, are lexically and syntactically simpler, and follow a more conventional style.
\citet{koppel-ordan-2011-translationese} show that classifiers can easily distinguish between original and translated text, and even determine the source language of the translated text. \citet{kong-macken-2025-decoding} show that this still holds true with LLMs, as distinguishing human from LLM-translated text works almost perfectly.
\citet{vanmassenhove-etal-2021-machine} show that machine translation systems tend to simplify language, losing lexical and morphological variety, 
but that Transformer-based models preserve more diversity than earlier systems.
Similarly, \citet{kong-macken-2025-decoding} find that LLMs generate more varied text than specialized machine translation systems, yet their outputs remain clearly distinguishable from human-written text.
\citet{bizzoni-etal-2020-human} compare human and machine translationese, finding that while human translations have similar properties across modalities (written versus spoken), machine translations exhibit fundamentally different patterns. 
\citet{li-etal-2025-lost} let annotators mark unnatural parts of translations, distinguishing between rigid sentence structures and literal word choices. They find that LLM outputs still contain much translationese, and trace this problem to translationese in the training data. They propose a polishing step where the model revises its own translation to reduce this effect, while prompts for natural style do not consistently help. 

Other studies have examined idiom translation as a common case of overly literal translation.
\citet{fadaee-etal-2018-examining} show that systems often translate idioms word by word, leading to semantic errors. 
\citet{dankers-etal-2022-transformer} also find that models tend to treat idioms compositionally, i.e., as literal phrases. When they recognize idioms as non-compositional units, interactions between the idiom's parts and the surrounding context decrease.
We annotate ``idioms'' as one category of error in our dataset to enable further research into this phenomenon.

\paragraph{English-to-Swedish translationese} \citet{gellerstam86} gives the classic example that English often uses an adjective together with a generic noun (e.g.\ \say{an [ADJ] thing}), whereas Swedish prefers to use a pronominal construction (\say{something [ADJ]}) instead. In translations from English, such noun constructions therefore appear more frequently than in original Swedish texts---e.g., \say{a silly thing happened} may be literally translated as \textit{en fånig sak hände} instead of the more idiomatic \textit{något fånigt hände} (\say{something silly happened}).
\citet{katourgi2022svenskan} documents other typical features of translation-influenced Swedish: Participial forms are less common than in English, e.g., \textit{Ta en bild på dig själv \underline{tittandes in} i kameran} (\say{Take a picture of yourself looking into the camera}) versus the more idiomatic \textit{\underline{när du tittar} mot kameran} (\say{\ldots when you look at the camera}). Also, noun phrases in predicative expressions often use the article when it should be omitted, e.g., \textit{Jag är \underline{en} översättare} $\rightarrow$ \textit{Jag är översättare} (\say{I am a translator} $\rightarrow$ \say{I am translator}). 
\citet{ahrenberg-2021-translation} analyzes adjective usage in English–Swedish translation and reports systematic distributional differences: human translators are more likely to restructure phrases, whereas systems tend to follow the source text more closely. \citet{ahrenberg-2017-comparing} compares a human and a machine translation of an article, reporting similar findings: the system adheres closely to the source, while the human employs strategies such as word reordering and sentence splitting, resulting in a longer text with a slightly higher type–token ratio.

\paragraph{Error tags for translations} 

For our dataset and analysis, we developed a custom error-tagging system. The tags were created through an iterative process in order to address the specific linguistic issues observed in our data. A related framework for error classification and assessment is the Multidimensional Quality Metrics \citep[MQM;][]{lommel-etal-2024-multi}.
MQM is currently regarded as a standard framework for analytic Translation Quality Evaluation. It consists of two central components: an \textit{error typology}, which provides a hierarchical classification of errors, and a \textit{scoring model}, which defines how identified errors are quantified (with different approaches depending on sample size). 
The MQM error typology is organized into seven main categories, with each one of them containing subcategories and subtypes of those. 
The scoring model is a combined method, process, and formula designed to derive overall quality scores from identified errors, either in calibrated or non-calibrated settings. Typically, the evaluation follows guidelines and specifications defined for a particular task or customer. Identified errors receive a quality score through assigned penalty points or weights, which are then aggregated in a record or scorecard.
While the MQM tag set has some overlap with our custom tag system, it also differs in important ways. In particular, it presents limitations for our analysis of idiomatic language use, where we need more fine-grained tags. The relationship between our tag system and the MQM tags is discussed in more detail in Appendix~\ref{app:mqm}.

\section{Dataset Construction}
\label{sec:dataset_construction}

Our dataset consists of 600~sentences from the English part of OpenSubtitles~\citelanguageresource{lison-tiedemann-2016-opensubtitles2016}, a dataset consisting predominantly of spoken dialogue. Sentences were translated to Swedish with OPUS-MT~\citep{tiedemann-thottingal-2020-opus} as an example for a neural (but not LLM-based) translation system, and {GPT-5}~\citep{openai2025gpt5} as an example for a recent LLM.
For each sentence, we provide error tags for each of the machine translations, an alternative translation produced by a human annotator, a contextual explanation, and a problem and solution description.

\subsection{Annotation Process}

The dataset was created and revised by two cognitive science students, both native Swedish speakers with basic linguistic training. The main annotator sampled random sections of the source document, collecting sentence pairs where OPUS translations showed signs of translationese.
For each pair, they added a brief context description, outlined the problem and its solution, and assigned up to three error tags (see Section~\ref{sec:error_tags}). 
The annotator then proposed more idiomatic alternative translations based on intuition and supported by dictionaries.
A third translation by GPT-5 was added and evaluated against the human alternative by the second annotator.
GPT-5 translations judged equally good or better were marked and supplemented with comments on their strengths or possible improvements. 
GPT-5 translations containing errors were tagged using the same scheme as the OPUS translations. 
The process was repeated several times for quality control, including the removal or replacement of problematic or duplicate entries.

\subsection{Error Tags}
\label{sec:error_tags}

Each phrase in the dataset is annotated with up to three tags to capture the types of issues encountered in the OPUS translation.
For the GPT-5 translation, we mark if the translation is acceptable or even improved in comparison to the human one.
We introduce three tags indicating \textbf{major errors}: 
\textit{Grammar}~(GR) is used for grammatical or syntactical errors, \textit{missing}~(SAK) for missing or not translated parts and \textit{incorrect}~(LF) for sentences containing words that are incorrect in the given context. \textit{Loss of meaning}~(BET) is used in combination with other tags and marks sentences where errors are critical enough to cause significant loss of the original phrase's meaning. \textit{Additional information}~ (ADD) is used when there are unnecessary added words in the translation. Multiple tags are used if there is need to capture multiple errors or connections between issues, e.g. if missing words cause loss of meaning. 
Two \textbf{minor error} tags mark up less critical, but still significant issues affecting interpretation. \textit{Semantic}~(SEM) indicates subtler changes in meaning that come with a risk of misinterpretation. \textit{Lexical preference}~(PR) is used for translations containing less normative, inappropriate or unnatural words from a fluent speaker's perspective. 
Finally, we use three 
\textbf{descriptive tags} to indicate the presence of certain types of language that are known to cause issues in machine translations: an \textit{idiom}~(ID) tag for idioms, a \textit{slang}~(SL) tag for informal language, a \textit{style}~(ST) tag for when the context requires domain-specific language and a \textit{direct translation}~ (DIR) tag for when the translation is a noticeable direct translation from the source language.
The descriptive tags are used to highlight causes of errors, combined with other tags that specify the issues that are connected to the original phrases of either category, e.g.\ idioms or slang expressions losing their meaning because of literal translation, or domain-specific terms not being applied where they would be preferrable.

\section{Dataset Analysis}
\label{sec:dataset_analysis}

Table~\ref{tab:tok-stats} shows basic token statistics for the Swedish translations from different sources, computed on the lower-cased text with punctuation marks removed.  The human translations are longest on average, both in terms of characters and (whitespace-separated) word tokens, and show the highest type--token ratio, which is in line with previous analyses of translationese \citep[e.g.,][]{ahrenberg-2017-comparing}.  The GPT-5 translations are closer to the human translations than the ones provided by OPUS-MT.

\begin{figure}
    \centering
    \begin{adjustbox}{width=\linewidth}
        \input{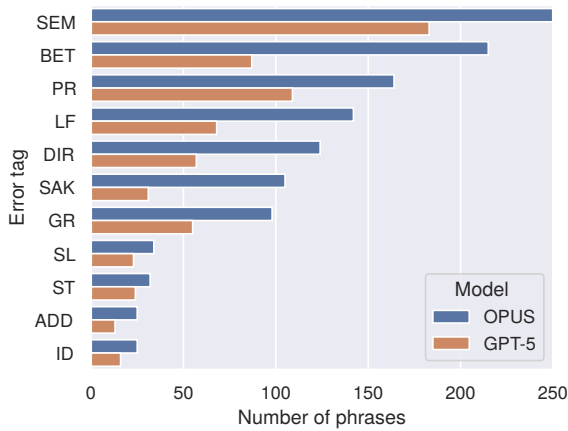}
    \end{adjustbox}
    \vspace{-15pt}
    \caption{Error tag distribution for the Swedish translations; cf.\ Sec.\,\ref{sec:error_tags} for an explanation of tags.}
    \label{fig:tag-stats}
\end{figure}

Figure~\ref{fig:tag-stats} shows the counts of each error tag by translation model.  The ``minor'' error tags SEM and PR are the most common, followed by loss of meaning~(BET) or incorrectly-translated words or phrases~(LF).  The GPT-5 translations consistently have fewer problems than the OPUS~translations, with the largest error reductions observed for PR and SAK, indicating that GPT-5 is much better at generating more idiomatic words or phrasings, and less likely to omit words entirely in the translation.

Comparing annotator judgments for GPT-5 versus the human translator, we find that the GPT-5~translation is equally acceptable in ca.\ 40\% of cases, and even judged to be an improvement over the human translation for 36~examples (6\%).

\begin{table}[]
    \centering
    \begin{adjustbox}{width=\linewidth}
    \begin{tabular}{lrrr}
    \toprule
    & \textbf{OPUS} & \textbf{Human} & \textbf{GPT-5} \\
    \midrule
    \textbf{\# word tokens}        & 3,860  & 3,971 & 3,913 \\
    \textbf{\# characters}         & 16,070 & 17,208 & 16,854 \\
    \textbf{Type--token ratio}  & 33.03  & 35.00 & 34.83 \\
    \bottomrule
    \end{tabular}
    \end{adjustbox}
    \caption{Basic statistics for the Swedish translations}
    \label{tab:tok-stats}
\end{table}

\subsection{Causes of Minor Errors}

\paragraph{Semantic shift (SEM)}
The most common error tag in our dataset is SEM, indicating that the situational or emotional meaning of the phrase is changed in a way that leaves room for misunderstanding.  For example, the phrase in~\pxref{ex:rest} is used in a context where the addressee is exhausted but doesn't go to bed, and therefore is told to rest:
\begin{subexamples}[preamble={You take rest\vspace{-.5em}}]\label{ex:rest}
    \item Ta det lugnt \expostamble{`take it easy' (SEM)}\label{ex:rest-opus}
    \item Du ska vila dig \expostamble{`you should/will rest' (SEM)}\label{ex:rest-gpt}
\end{subexamples}
The OPUS translation~\pxref{ex:rest-opus} captures that the person needs to calm down in some manner, but does not communicate the form of rest; the GPT-5~translation~\pxref{ex:rest-gpt} carries a more aggressive connotation than ``care.''  An idiomatic translation would be \textit{Ta och vila}, roughly translating to `take some rest.'

Metaphoric language is a common cause of this error type.  The phrase in~\pxref{ex:lookclean} is used in the context of policemen offering help to an alleged criminal, and translated identically by OPUS and GPT-5:
\begin{example}[preamble={We could make you \hla{look clean}}]\label{ex:lookclean}
    Vi kan få dig att \hla{se ren ut} \expostamble{(SL, SEM)}
\end{example}
The translation here carries a literal meaning and could also be interpreted as drug-related. To express this is in a more idiomatic way, the word \textit{fläckfri} `spotless' is more precise than  \textit{ren}~`clean'.

\begin{subexamples}[preamble={They just \hla{fade in}\vspace{-.5em}}]\label{ex:fadein}
    \item De bara \hla{bleknar in} \expostamble{(SEM)}\label{ex:fadein-opus}
    \item De bara \hla{tonar in} \expostamble{`they just tune in'\\ (LF, BET)}\label{ex:fadein-gpt}
\end{subexamples}
Example~\pxref{ex:fadein} is used in the context of black people ``fading in'' in society.  The OPUS translation~\pxref{ex:fadein-opus} uses the word \textit{bleknar} `to become pale', which is possible to understand, but not a commonly accepted usage of the word.  GPT-5 produces a worse translation~\pxref{ex:fadein-gpt} in this case, using the word \textit{tonar} that can be used in reference to ``fading in music'' or to ``engage in something,'' which communicates the opposite of the original phrase, and thus gets annotated with different error tags in our dataset. A more idiomatic, and still metaphorical, phrasing that is appropriate in the given context would be \textit{De bara \hla{smälter in}}, literally `they just melt in.'

\paragraph{Lexical preference (PR)}
This tag indicates translations that communicate the meaning but sound unnatural to a native speaker.
For example, while \textit{gåva} in Swedish does mean `gift,' it has a broader and more formal meaning than in English, and sounds unnatural in a more casual context:
\begin{example}[preamble={Simone, your \hla{gift}}]\label{ex:gift}
    Simone, din \hla{gåva} \expostamble{(PR)}
\end{example}
A more appropriate choice here is \textit{present}~`present', which has a less formal connotation in Swedish.

Translations can also be too informal, as in~\pxref{ex:murder}, where the English phrase communicates the seriousness of a crime in a police investigation:
\begin{example}[preamble={We're \hla{talking} murder, here}]\label{ex:murder}
    Vi \hla{pratar} om mord \expostamble{(SAK, PR)\\`We're speaking of murder here'}
\end{example}
Both OPUS and GPT-5 pick the word \textit{pratar}~`speak', which is not wrong, but makes the information sound less serious than it is.  In a context like this, \textit{talar}~'talk' would be a more formal choice, thus conveying the gravity of the situation better.

\begin{example}[preamble={Poor little \hla{thing}}]\label{ex:littlething}
    Stackars lilla \hla{sak} \expostamble{(DIR, SEM, PR)}
\end{example}
The phrase in~\pxref{ex:littlething} is said to comfort another person and sympathize with them; the direct, literal, translation uses the word \textit{sak}~`thing', which in Swedish is only applicable when referring to items, not persons. To call someone an \say{item} is not something one would do in a serious and emotional situation, so we use SEM to highlight that this sentence could be interpreted as unserious or rude.  A common and lexically preferred phrase to would be \textit{Din stackare}, roughly translating to `you poor one/person.'

\subsection{Causes of Major Errors}

\paragraph{Loss of meaning (BET)}
The most common ``major'' problem we observe is that the phrase's meaning gets lost in the translation. In~\pxref{ex:knark}, the word ``dope'' is used as an intensifier synonymously with ``cool'' or ``awesome,'' to describe a watch, whereas OPUS literally translates it to ``drug watch'':
\begin{example}[preamble={This \hla{dope} watch}]\label{ex:knark}
    Den här \hla{knark}klockan \expostamble{`this drug watch'\\(SL, DIR, BET)}
\end{example}
A more correct translation would be \textit{den här \hla{fräna} klockan}, using the adjective \textit{frän}~`cool, stylish' that is commonly used in such contexts.

Idioms~(ID) are a common source of this type of error, as in Example~\pxref{ex:bones}, which talks about the structural properties of a house:
\begin{subexamples}[preamble={It's got good \hla{bones}\vspace{-.5em}}]\label{ex:bones}
    \item Den har bra \hla{ben} \expostamble{(ID, DIR, BET)}\label{ex:bones-opus}
    \item Den har bra \hla{grundförutsättningar} \expostamble{`it has good basic prerequisites' (ID, SEM)}\label{ex:bones-gpt}
\end{subexamples}
The literal OPUS translation~\pxref{ex:bones-opus} loses the idiomatic meaning in this context.  GPT-5 correctly captures this meaning in~\pxref{ex:bones-gpt}, though the phrasing causes a subtle semantic shift due to the idiomaticity being lost.  An idiomatic alternative that is applicable in the context of housing would be \textit{Den har bra stomme}~`it has good framing.'

\paragraph{Incorrect word choice (LF)}
Similar to~\pxref{ex:bones}, Example~\pxref{ex:oldbone} contains an idiom where the OPUS translation~\pxref{ex:oldbone-opus} loses the meaning by being overly literal.  It also translates the word `pick' as \textit{välja}~`choose', which is not the correct choice in this context, and therefore gets annotated with the LF~tag:
\begin{subexamples}[preamble={I had \hla{an old bone} \hlb{to pick with you}\vspace{-.5em}}]\label{ex:oldbone}
    \item Jag hade \hla{ett gammalt ben} \hlb{att välja med dig} \expostamble{`I had an old bone to choose with you' (ID, LF, BET)}\label{ex:oldbone-opus}
    \item Jag hade ett gammalt horn i sidan på dig \expostamble{`I had an old horn in your side'\\(ID, LF, BET)}\label{ex:oldbone-gpt}
\end{subexamples}
The GPT-5 translation~\pxref{ex:oldbone-gpt} attempts to use the Swedish idiom \textit{ett horn i sidan}, related to the English `a thorn in the side,' but this is neither fitting nor used correctly here.  A Swedish idiom that would fit better here is \textit{Jag hade en oplocked gås med dig}, literally `I had an unplucked goose with you,' used in the same way as the original phrase.

\begin{example}[preamble={He's a \hla{degenerate} gambler}]\label{ex:gambler}
    Han är en \hla{degenererad} spelare \expostamble{`he is a degenerat\underline{ed} player/gambler' (LF, BET, PR)}
\end{example}
Example~\pxref{ex:gambler} mistranslates `degenerate' by applying the closely related adjective \textit{degenererad} `degenerated', thus also losing the meaning.  To capture the intent of the original utterance, we could use the phrase \textit{Han är en \hla{nedgången} speltorsk}, roughly `he is a worn-out gambling addict.'

\paragraph{Grammatical errors (GR)}
Example~\pxref{ex:alla} illustrates a case where both OPUS and GPT-5 introduce grammatical errors into the translation by attaching `all' to the object rather than the subject:
\begin{subexamples}[preamble={We \hla{all} miss \hlb{you}\vspace{-.5em}}]\label{ex:alla}
    \item<*> Vi saknar \hlb{dig} \hla{alla} \expostamble{(GR)}\label{ex:alla-opus}
    \item<*> Vi saknar \hlb{dig} \hla{allihop} \expostamble{(GR)}\label{ex:alla-gpt}
\end{subexamples}
Both translations rather suggest the meaning `We miss you all,' but are grammatically incorrect as \textit{alla} `all' and \textit{allihop} `all (together)' are plural pronouns, where \textit{dig} `you' is exclusively singular.  The correct translation should follow the same word order as in English, i.e.\ \textit{Vi \hla{alla} saknar \hlb{dig}}.

We also observe examples of incorrect article use that were described by \citet{katourgi2022svenskan}, e.g.\ in predicative constructions describing a person's occupation, where the article \textit{en} needs to be omitted for a grammatically correct translation:
\begin{example}[preamble={You're \hla{a} detective}]\label{ex:detective}
    *\,Du är \hla{en} detektiv \expostamble{(GR)}
\end{example}

\begin{figure*}[t]
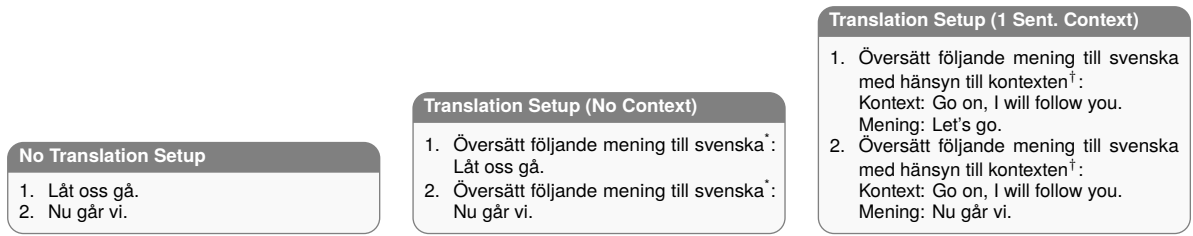

\centering
\begin{tabularx}{\linewidth}{XXX}
\begin{tcolorbox}[title=No Translation Setup, boxsep=2pt, left=2pt, right=2pt, top=2pt, bottom=2pt]
\begin{enumerate}[leftmargin=*, nosep, topsep=0pt]
\item Låt oss gå.
\item Nu går vi.
\end{enumerate}
\end{tcolorbox}

&
\begin{tcolorbox}[title=Translation Setup (No Context), boxsep=2pt, left=2pt, right=2pt, top=2pt, bottom=2pt]
\begin{enumerate}[leftmargin=*, nosep, topsep=0pt]
\item Översätt följande mening till svenska\textsuperscript{*}: Låt oss gå.
\item Översätt följande mening till svenska\textsuperscript{*}: Nu går vi.
\end{enumerate}
\end{tcolorbox}

&
\begin{tcolorbox}[title=Translation Setup (1 Sent.\ Context), boxsep=2pt, left=2pt, right=2pt, top=2pt, bottom=2pt]
\begin{enumerate}[leftmargin=*, nosep, topsep=0pt]
\item Översätt följande mening till svenska med hänsyn till kontexten\textsuperscript{$\dagger$}:\par
Kontext: Go on, I will follow you.\par
Mening: Let's go.

\item Översätt följande mening till svenska med hänsyn till kontexten\textsuperscript{$\dagger$}:\par
Kontext: Go on, I will follow you.\par
Mening: Nu går vi.
\end{enumerate}
\end{tcolorbox}
\end{tabularx}

\caption{Prompting setups. Each box shows a \textbf{minimal pair}: Sentence~1 is a translationese variant, sentence~2 is an idiomatic variant of a translation of the same English sentence. We compute the perplexity of each variant to determine which one the model prefers.  Translations of text in the figure: \textsuperscript{*}\textit{Translate the following sentence to Swedish.} \textsuperscript{$\dagger$}\textit{Translate the following sentence to Swedish, considering the context.}}
\label{fig:prompt_setups}
\end{figure*}

\paragraph{Missing words (SAK)}
This category is much more common in the OPUS~translations than in GPT-5.  In many cases, the missing words do not affect the translation significantly, e.g.\ when a sentence-initial `Come on' or `Listen, ...' is omitted, or in Example~\pxref{ex:hands}, where the verb is omitted from the imperative clause, causing no errors or misunderstandings, but still differs from the original.
\begin{example}[preamble={\hlb{Put your} hands on your head!}]\label{ex:hands}
    Händerna på huvudet! \expostamble{`Hands on your head!' (SAK)}
\end{example}
\begin{subexamples}[preamble={\hlb{Why} don't you \hla{go get some sleep}?\vspace{-.5em}}]\label{ex:gosleep}
    \item \hla{Gå och lägg dig}. \expostamble{`Go to sleep'\\ (SAK, SEM)}\label{ex:gosleep-opus}
    \item \hlb{Varför} \hla{går} du inte \hla{och sover lite}? \expostamble{`Why don't you go and get some sleep?' (SEM)}\label{ex:gosleep-gpt}
\end{subexamples}
Example~\pxref{ex:gosleep} is used to suggest, in a caring way, that a person should go to bed.  The OPUS translation~\pxref{ex:gosleep-opus} is missing a large part of the source sentence and reads more like an order. 
GPT-5 produces a direct translation~\pxref{ex:gosleep-gpt} of the original; however, in Swedish, this shifts the phrase's tone from caring to questioning.  A more idiomatic phrasing would be \textit{Ska inte du gå och försöka sova lite?}, roughly `Shouldn't you go and try sleeping a little?'

\subsection{Domain-Specific Language}
Some causes of errors that we annotate with their own tag have already been mentioned above, such as slang words (SL; Ex.~\ref{ex:knark}) and idioms (ID; Ex.~\ref{ex:bones} and~\ref{ex:oldbone}).  Domain-specific language~(ST), i.e.\ phrases that use terminology specific to a certain field or domain, are another common error source in our dataset.
An example of this is the legal domain, which often uses very specific terminology:
\begin{example}[preamble={\hla{Not guilty} \hlb{on all counts}}]\label{ex:notguilty}
    \hla{Inte skyldig} \hlb{på alla punkter} \expostamble{(ST, PR)}
\end{example}
Here, the OPUS~translation is again very literal and understandable, although better law-specific terminology exists: The human translation in our dataset is \textit{\hla{Oskyldig} \hlb{på samtliga åtalspunkter}}, using \textit{oskyldig} `innocent' instead of the literal `not guilty,' the more formal and therefore lexically-preferred variation \textit{samtliga} `all,' and the legal term \textit{åtalspunkter}, literally `counts of indictment.'

Example~\pxref{ex:counsel} shows a subtle shift in meaning from the occupational term of `counselling':
\begin{subexamples}[preamble={I \hla{counsel} people with AIDS}]\label{ex:counsel}
    \item Jag \hla{ger råd till} människor med AIDS \expostamble{\\`I {give advice to} people with AIDS' \\(ST, SEM)}\label{ex:counsel-opus}
    \item Jag \hla{ger stöd till} personer med AIDS \expostamble{\\`I {give support to} persons with AIDS' \\(ST, SEM)}\label{ex:counsel-gpt}
\end{subexamples}
Both translations capture the practical aspects of the job, but leave out the professional part of being a counselor.  The correct Swedish term here is \textit{kurator} `counselor,' but this cannot be conjugated into a verb, which means that a nominal construction must be used, e.g.\ \textit{Jag \hla{är kurator åt} människor med AIDS} `I {am a counselor for} people with AIDS.'

\section{Experiments}
\label{sec:experiments}

\begin{table*}[h!]
\centering
\resizebox{0.85\linewidth}{!}{%
\begin{tabulary}{\linewidth}{lRR>{\quad}RR>{\quad}RR>{\quad}RR}
\toprule
\textbf{Model} & \multicolumn{2}{w{c}{7em}}{\textbf{Human$>$OPUS}} & \multicolumn{2}{c}{\textbf{Human$>$GPT-5}} & \multicolumn{2}{c}{\textbf{SGB$_{all}$}} & \multicolumn{2}{c}{\textbf{SGB$_{filtered}$}} \\
\cmidrule(lr){2-3} \cmidrule(lr){4-5} \cmidrule(lr){6-7} \cmidrule(lr){8-9}
 & Acc. & $\Delta$LP & Acc. & $\Delta$LP & Acc. & $\Delta$LP & Acc. & $\Delta$LP  \\
\midrule
LLaMA-3-8B & 47.83 & 1.83 & 40.33 & 0.91 & 38.57 & -3.01 & 56.71 & 7.13 \\
\vspace{2pt}
LLaMA-3-8B-it & 49.83 & 2.01 & 42.50 & 1.17 & 40.33 & -2.65 & 58.95 & 8.47 \\
ai.se-LLaMA-8B & 56.67 & 5.17 & 44.50 & 2.36 & 49.01 & 3.67 & 81.34 & 16.56 \\
\vspace{2pt}
ai.se-LLaMA-8B-it & 52.83 & 3.36 & 43.50 & 1.47 & 34.12 & -5.80 & 58.95 & 5.60 \\
\midrule
EuroLLM-1.7B & 48.33 & 1.09 & 41.17 & 0.83 & 41.46 & -2.01 & 58.06 & 4.70 \\
\vspace{2pt}
EuroLLM-1.7B-it & 49.17 & 2.00 & 40.83 & 1.05 & 41.26 & -2.34 & 57.41 & 4.88 \\
EuroLLM-9B & 49.67 & 2.64 & 43.00 & 1.19 & 42.19 & -0.80 & 61.29 & 5.41 \\
\vspace{2pt}
EuroLLM-9B-it & 51.00 & nan & 40.83 & nan & 44.26 & nan & 65.35 & nan \\
\midrule
Gemma-270M & 47.33 & 0.95 & 45.83 & 3.16 & 42.50 & -0.13 & 56.33 & 6.22 \\
\vspace{2pt}
Gemma-270M-it & 49.50 & 0.61 & 44.00 & 2.40 & 40.95 & -1.15 & 49.29 & 4.49 \\
Gemma-1B & 49.67 & 2.99 & 47.33 & 3.93 & 42.91 & 0.11 & 63.38 & 7.64 \\
\vspace{2pt}
Gemma-1B-it & 46.00 & 0.88 & 43.50 & 2.80 & 40.12 & -2.37 & 51.40 & 2.43 \\
Gemma-4B & 52.50 & 4.03 & 46.33 & 3.83 & 43.22 & -0.36 & 66.90 & 7.86 \\
\vspace{2pt}
Gemma-4B-it & 54.83 & 6.49 & 46.00 & 4.49 & 38.77 & -2.29 & 56.33 & 3.91 \\
Gemma-12B & 55.33 & 5.54 & 45.00 & 4.39 & 43.64 & 0.39 & 71.12 & 8.79 \\
\vspace{2pt}
Gemma-12B-it & 58.33 & 8.31 & 46.33 & 4.09 & 40.53 & -1.94 & 59.85 & 6.89 \\
\bottomrule
\end{tabulary}}
\caption{Preferences for prompts \textit{without} translation context for human alternatives over OPUS or GPT translations. We compare to the datasets by \citet{anonymous2025} derived from \citet{katourgi2022svenskan} (\textit{SGB}).}
\label{tab:preferences}
\end{table*}

\paragraph{Prompting setups} We evaluate models on our dataset in a minimal pairs setup (translationese versus alternative) to probe intrinsic preferences for idiomatic language. 
We use two prompting setups to examine how contextual and task framing affect their preferences. Each setup compares two alternatives: a \textit{machine translation} and the \textit{human alternative} version of a sentence. 
In the \textbf{no translation context} setup, the model is simply presented with the Swedish sentence, 
allowing us to assess general preference without translation instruction. 
In the \textbf{with translation context} setup, the model is instructed to translate the English source sentence into Swedish. 
We test seven variants: one where the model receives only the target sentence, 
and six where the prompt even includes 1--10 preceding English sentences as contextual information. 
The an example for the prompting setups is illustrated in Figure~\ref{fig:prompt_setups}.

\paragraph{Models}
We evaluate on language models across different scales and language coverage. 
AI Sweden LLaMA-3 8B\footnote{\url{https://huggingface.co/AI-Sweden-Models/Llama-3-8B}} is a continued pre-training of LLaMA-3 8B \citep{grattafiori2024llama3herdmodels} on Scandinavian language data. To quantify the impact of this adaptation, we also include the original \textbf{LLaMA-3} 8B as a baseline. EuroLLM-1.7B and 9B \citep{martins2024eurollmmultilinguallanguagemodels} are open multilingual language models that include Swedish in their pre-training corpus. Finally, we include the Gemma-3 models \citep{gemmateam2025gemma3technicalreport} (270M, 1B, 4B, and 12B) as multilingual models across a wider size range. We include both base models and instruction tunes (suffix \textit{-it}).

\paragraph{Metrics} 
A direct comparison of log-likelihoods across sentences is problematic as the alternatives often differ in length. To account for this, we use the length-normalized mean log probability ($\MeanLP$), defined as 
\begin{equation}\label{eq:meanlp}
\MeanLP(x) = \frac{1}{|x|} \sum_{i=1}^{|x|} \log p(w_i \mid w_{<i})
\end{equation}
where $x = (w_1, \dots, w_{|x|})$ is the sentence and $|x|$ is its length. 
We use it for two metrics: The 
\textbf{Accuracy}, i.e., the percentage of examples where the human alternative receives a higher probability than the OPUS or GPT sentence, 
capturing \textit{how often} the model prefers the human variant,
and \boldmath$\Delta$\textbf{LP}, i.e., the average relative difference (\%) between the probabilities of the OPUS or GPT and human variants across the dataset. This metric reflects the \textit{magnitude} of the model’s preference. A negative value indicates a stronger preference for the translationese variant.

\paragraph{Reference dataset} We compare our results to the dataset introduced by \citet{anonymous2025}, based on the book \textit{Svenskan går bananer} \citep{katourgi2022svenskan}, which contrasts Translationese sentences with idiomatic alternatives suggested by the author. Following \citet{anonymous2025}, we evaluate two setups: (1) SGB\textsubscript{all}, which includes all sentence pairs from the book, and (2) SGB\textsubscript{filtered}, a version filtered to include only pairs where (a) both sentences are of the same length, since sentence lengths vary considerably in the dataset, and (b) human annotators agreed that the alternative is clearly better than the translationese version. We use this dataset to compare whether similar patterns, such as model rankings, hold using their dataset and in ours.

\section{Results}

\begin{table*}[h!]
\centering
\begin{subtable}[t]{\textwidth}
\begin{adjustbox}{width=\textwidth}
\centering
\begin{tabular}{lrrrrrrrrrrrrrr}
\toprule
\textbf{Model} & \multicolumn{2}{c}{\textbf{0 Sent.}} & \multicolumn{2}{c}{\textbf{1 Sent.}} & \multicolumn{2}{c}{\textbf{2 Sent.}} & \multicolumn{2}{c}{\textbf{3 Sent.}} & \multicolumn{2}{c}{\textbf{4 Sent.}} & \multicolumn{2}{c}{\textbf{5 Sent.}} & \multicolumn{2}{c}{\textbf{10 Sent.}} \\
\cmidrule(lr){2-3} \cmidrule(lr){4-5} \cmidrule(lr){6-7} \cmidrule(lr){8-9} \cmidrule(lr){10-11} \cmidrule(lr){12-13} \cmidrule(lr){14-15}
 & Acc. & $\Delta$LP & Acc. & $\Delta$LP & Acc. & $\Delta$LP & Acc. & $\Delta$LP & Acc. & $\Delta$LP & Acc. & $\Delta$LP & Acc. & $\Delta$LP \\
\midrule
LLaMA-3-8B & 40.50 & -1.75 & 39.33 & -1.11 & 40.50 & -0.87 & 41.50 & -0.68 & 42.50 & -0.61 & \textbf{42.67} & -0.51 & 42.17 & -0.32\\
\vspace{2pt} LLaMA-3-8B-it & 39.33 & -2.28 & 42.00 & -1.03 & 42.83 & -0.77 & 44.17 & -0.66 & 44.00 & -0.58 & 43.83 & -0.53 & \textbf{44.33} & -0.32\\
ai.se-LLaMA-8B & 43.00 & -1.58 & 49.33 & -0.10 & 48.50 & 0.03 & 49.67 & 0.11 & 51.17 & 0.12 & \textbf{51.33} & 0.12 & 50.67 & 0.08\\
\vspace{2pt} ai.se-LLaMA-8B-it & 44.00 & -1.39 & 44.67 & -0.95 & 45.33 & -0.68 & 45.50 & -0.52 & 46.17 & -0.44 & 46.50 & -0.37 & \textbf{46.83} & -0.21\\
\midrule
EuroLLM-1.7B & 33.67 & -3.09 & 35.17 & -1.69 & 37.33 & -1.31 & 37.17 & -1.05 & 37.00 & -0.94 & 36.83 & -0.83 & \textbf{38.17} & -0.57\\
\vspace{2pt} EuroLLM-1.7B-it & 35.50 & -2.77 & 38.50 & -1.58 & \textbf{40.50} & -1.21 & 39.17 & -0.96 & 40.00 & -0.85 & 39.83 & -0.77 & 40.00 & -0.53\\
EuroLLM-9B & 41.33 & -1.43 & 41.00 & -0.90 & 42.00 & -0.61 & 44.83 & -0.40 & 42.83 & -0.34 & \textbf{45.67} & -0.26 & 45.50 & -0.15\\
\vspace{2pt} EuroLLM-9B-it & 38.83 & -1.90 & 42.83 & -0.90 & 43.17 & -0.64 & 45.50 & -0.39 & 44.67 & -0.31 & 44.00 & -0.23 & \textbf{47.00} & -0.11\\
\midrule
Gemma-270M & 35.17 & -2.42 & 35.00 & -1.82 & 34.50 & -1.58 & 35.00 & -1.35 & \textbf{35.50} & -1.23 & \textbf{35.50} & -1.11 & \textbf{35.50} & -0.79\\
\vspace{2pt} Gemma-270M-it & 32.17 & -3.73 & 32.00 & -2.46 & 32.17 & -2.15 & 33.33 & -1.87 & 33.00 & -1.65 & 34.00 & -1.52 & \textbf{35.17} & -1.07\\
Gemma-1B & 36.50 & -2.41 & 38.67 & -1.19 & 39.83 & -0.91 & 40.17 & -0.77 & 39.50 & -0.69 & 40.00 & -0.62 & \textbf{42.17} & -0.43\\
\vspace{2pt} Gemma-1B-it & 33.00 & -3.60 & 35.83 & -2.04 & 36.33 & -1.47 & \textbf{38.33} & -1.22 & 36.83 & -1.06 & 37.83 & -0.93 & 38.00 & -0.70\\
Gemma-4B & 40.17 & -1.38 & 41.50 & -0.90 & 42.50 & -0.69 & 42.83 & -0.52 & 43.50 & -0.44 & \textbf{44.67} & -0.39 & \textbf{44.67} & -0.27\\
\vspace{2pt} Gemma-4B-it & 41.50 & -1.53 & 43.00 & -0.76 & 45.00 & -0.58 & 46.33 & -0.43 & 45.83 & -0.32 & \textbf{46.83} & -0.32 & 44.33 & -0.24\\
Gemma-12B & 45.50 & -0.61 & 44.50 & -0.47 & 46.33 & -0.27 & 46.50 & -0.19 & \textbf{46.83} & -0.17 & 46.00 & -0.14 & 45.50 & -0.11\\
\vspace{2pt} Gemma-12B-it & 48.67 & 0.79 & 51.33 & 0.91 & 52.00 & 0.95 & 54.83 & 1.01 & 56.17 & 0.92 & \textbf{57.00} & 0.88 & 55.67 & 0.63\\
\bottomrule
\end{tabular}
\end{adjustbox}
\caption{Human $>$ OPUS}
\vspace{2pt}
\label{tab:pref_translation_opus}
\end{subtable}
\begin{subtable}[t]{\textwidth}
\begin{adjustbox}{width=\textwidth}
\centering
\begin{tabular}{lrrrrrrrrrrrrrr}
\toprule
\midrule
LLaMA-3-8B & \textbf{33.17} & -2.15 & 33.00 & -1.43 & 33.00 & -1.25 & 32.50 & -1.07 & 33.00 & -0.97 & 32.67 & -0.87 & \textbf{33.17} & -0.60\\
\vspace{2pt} LLaMA-3-8B-it & 30.67 & -3.12 & 31.67 & -1.72 & 32.17 & -1.46 & 32.83 & -1.31 & 32.83 & -1.21 & 31.67 & -1.11 & \textbf{34.00} & -0.74\\
ai.se-LLaMA-8B & \textbf{34.17} & -1.64 & 34.00 & -0.91 & 33.83 & -0.78 & 33.50 & -0.68 & 33.83 & -0.63 & \textbf{34.17} & -0.57 & \textbf{34.17} & -0.42\\
\vspace{2pt} ai.se-LLaMA-8B-it & 30.67 & -2.63 & 30.67 & -1.78 & 30.50 & -1.49 & 31.00 & -1.30 & 30.33 & -1.17 & 30.00 & -1.06 & \textbf{31.67} & -0.71\\
\midrule
EuroLLM-1.7B & 30.17 & -2.33 & 30.33 & -1.41 & 31.50 & -1.15 & 32.33 & -0.96 & 32.83 & -0.86 & 32.50 & -0.79 & \textbf{33.17} & -0.55\\
\vspace{2pt} EuroLLM-1.7B-it & 29.00 & -2.48 & 30.33 & -1.53 & 31.83 & -1.24 & 32.00 & -1.06 & \textbf{32.83} & -0.96 & 31.83 & -0.88 & \textbf{32.83} & -0.62\\
EuroLLM-9B & 25.83 & -3.04 & 25.17 & -1.98 & 26.50 & -1.62 & 26.83 & -1.37 & 27.17 & -1.23 & 26.67 & -1.09 & \textbf{28.33} & -0.75\\
\vspace{2pt} EuroLLM-9B-it & 24.17 & -3.73 & 25.00 & -2.20 & 26.00 & -1.82 & \textbf{27.83} & -1.52 & 26.50 & -1.34 & 26.50 & -1.18 & 27.67 & -0.79\\
\midrule
Gemma-270M & 34.00 & -1.13 & 33.83 & -0.92 & 34.83 & -0.76 & \textbf{36.33} & -0.63 & 34.33 & -0.62 & 34.67 & -0.54 & 35.50 & -0.38\\
\vspace{2pt} Gemma-270M-it & 36.00 & -1.62 & 33.33 & -1.20 & 36.67 & -1.02 & 36.83 & -0.89 & 36.17 & -0.77 & 36.17 & -0.69 & \textbf{37.33} & -0.48\\
Gemma-1B & 33.33 & -1.69 & 34.17 & -0.95 & \textbf{35.50} & -0.78 & 33.50 & -0.67 & 33.17 & -0.63 & 33.67 & -0.55 & 33.67 & -0.43\\
\vspace{2pt} Gemma-1B-it & 30.50 & -2.41 & 31.33 & -1.43 & 32.33 & -1.16 & 33.00 & -0.94 & 32.33 & -0.81 & 33.33 & -0.73 & \textbf{34.83} & -0.51\\
Gemma-4B & 31.67 & -1.82 & 32.00 & -1.25 & 33.17 & -1.05 & 32.83 & -0.89 & 33.83 & -0.79 & 33.83 & -0.71 & \textbf{34.50} & -0.51\\
\vspace{2pt} Gemma-4B-it & 26.33 & -3.70 & 28.17 & -2.19 & 29.17 & -1.84 & \textbf{30.50} & -1.58 & 30.00 & -1.43 & 29.17 & -1.29 & 30.17 & -0.87\\
Gemma-12B & 31.83 & -1.89 & 30.67 & -1.27 & 32.17 & -1.02 & 31.83 & -0.90 & 32.00 & -0.81 & 31.17 & -0.74 & \textbf{32.67} & -0.52\\
\vspace{2pt} Gemma-12B-it & 29.67 & -3.63 & 28.17 & -2.47 & 27.83 & -2.09 & 29.17 & -1.82 & 29.33 & -1.63 & 30.00 & -1.47 & \textbf{30.50} & -0.98\\
\bottomrule
\end{tabular}
\end{adjustbox}
\caption{Human $>$ GPT}
\vspace{2pt}
\label{tab:pref_translation_gpt}
\end{subtable}
\caption{Preferences for machine translations vs.\ human alternatives for prompts \textit{with} translation context, with 0--10 preceding sentences from the source document. Highest scores for each model are bold.}
\label{tab:with_transl_context}
\end{table*}

All models we evaluated show a bias toward the machine translated sentences, even for OPUS variants where the translationese wording is obvious. As shown in Tables~\ref{tab:preferences}, \ref{tab:pref_translation_opus}, and \ref{tab:pref_translation_gpt}, there are very few instances where models prefer the human translation over the machine translation in most examples.

\paragraph{Preferences without translation context} 

In Table~\ref{tab:preferences}, we see that the models often prefer the machine translated sample. The highest selection rate (accuracy) for the human alternative over OPUS is $58.33$ for the largest model, Gemma-12B-it. However, even though most models prefer the machine-translated sentences in the majority of cases, the $\Delta$LP values are positive in all settings except for the SGB\textsubscript{all} dataset (which contains strong length-based cues favoring the translationese samples). This suggests that when a model \textit{does} prefer the human alternative, it tends to assign it a higher likelihood than it does for machine-translated sentences when the opposite is true. 
Table~\ref{tab:preferences} also shows that the Human$>$OPUS setup exhibits some expected trends: scores increase with model size for both Gemma and EuroLLM families, and ai.se-LLaMa-8B gets higher scores than Llama-3-8B. This pattern is similar to that observed for SGB$_{filtered}$, although the latter has overall higher scores. There are only two exceptions to these trends: in the Human$>$OPUS setup, accuracy does not increase between Gemma-135M-it and Gemma-1B-it, and in the SGB\textsubscript{filtered} setup, the ai.se-LLaMa-8B-it does not outperform Llama-3-8B-it.\footnote{\citet{anonymous2025} also found that instruction tuning biases ai.se-LLaMa-8B towards translationese.}
The Human$>$GPT setup produces generally lower and more variable results than Human$>$OPUS, similar to those for SGB\textsubscript{all}, although slightly higher and with positive $\Delta$LP values. For these two datasets, scores do not consistently increase e.g.\ with model size. 
The lower scores in the Human$>$GPT setup align with our analysis in Section~\ref{sec:dataset_analysis}, where we found that GPT translations are often more acceptable, even if many still contain issues. Consequently, comparisons against GPT are a less reliable indicator of model capability than comparisons against OPUS.

\begin{figure*}
    \centering
    \begin{adjustbox}{width=\linewidth}
        \input{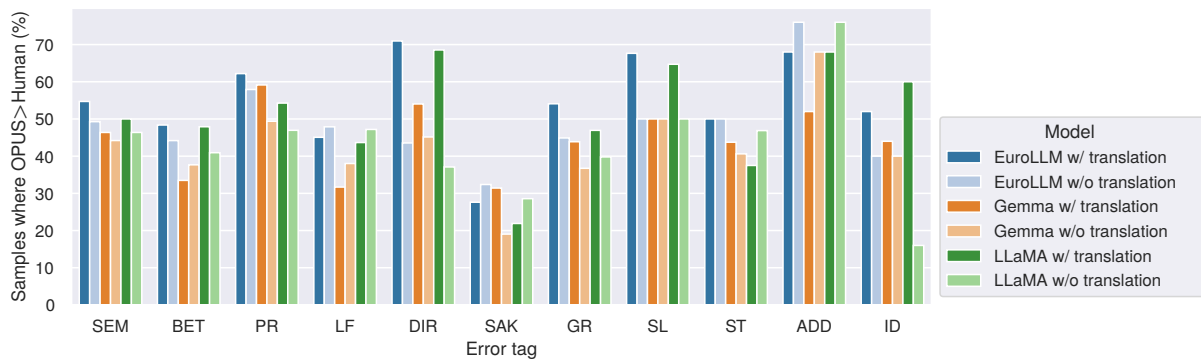}
    \end{adjustbox}
    \vspace{-12pt}
    \caption{Percentage of samples where models prefer the OPUS-translated sentence over the human alternative, by error tag. See Section~\ref{sec:error_tags} for explanations of error tags. }
    \label{fig:model_errors_tags}
\end{figure*}

\paragraph{With and without translation context}
Adding translation context could help the model interpret how an expression is intended, but it may also bias it toward preferring the more literal machine-translated sentence. 
Comparing the Human$>$OPUS setup in Table~\ref{tab:preferences} to the best values in Table~\ref{tab:pref_translation_opus}, we see that when \textit{no} translation context is provided, results are substantially better: accuracy is higher than even in the best translation setups, and $\Delta$LP values are more often positive. We thus see that the English source sentence steers the model toward the literal, translationese option, even with a substantial amount of context. 

\paragraph{Translation contexts, with varying context length}

As visible in Table~\ref{tab:pref_translation_opus}, adding more context often leads to a clear improvement: accuracy is \textit{always} higher with two or more preceding sentences than with none. 
This suggests that context helps the model better interpret the intended meaning and, as a result, favor the human translation in some cases. Five sentences of context is often the best overall setup, as indicated by the bolded scores in Table~\ref{tab:pref_translation_opus}, followed by ten sentences. This shows that a substantial amount of context generally helps better to avoid Translationese phrasings. 
Even so, very few models achieve positive $\Delta$LP values (only the base AI Sweden LLaMA for half of the prompts and the Gemma-12B-it model across all prompts). 
We also see a strong \say{bigger is better} trend for Human$>$OPUS.
For the Human$>$GPT values in Table~\ref{tab:pref_translation_gpt} however, the opposite holds true: larger models generally perform worse. There is often a consistent decrease in accuracy with increasing model size, both within the EuroLLM and Gemma families: As the models become more capable, their preference for the GPT-produced translations tends to \textit{increase}. A clear trend for Human$>$GPT is that the setup with 10 sentences is the best; the bolded values show that it is the best for 12 out of 16 models. There appears to be a number of pairs in the dataset where the model does prefer the human to the GPT-5 translation, but only with the help of substantial context. 

\paragraph{Error tags} 
We explore which types of errors are over- and underrepresented in cases where the OPUS translation is preferred over the human alternative. Specifically, we analyze the best-performing models from each family: AI Sweden LLaMA base, and the instruction-tuned versions of EuroLLM-9B and Gemma-12B. The analysis is carried out using the best-performing setups: (i) without a translation prompt and (ii) with a translation prompt including a 10-sentence context. 
As shown in Figure~\ref{fig:model_errors_tags}, direct translations~(DIR) stand out as strongly overrepresented when translation context is provided (54--71\% OPUS preference), a sharp increase from the no-context setting (37--45\%), suggesting that context may lead models to translate too literally. Slang~(SL) also leads to many errors, particularly with context (up to 68\% for EuroLLM), showing that such expressions are hard to translate idiomatically. Even the minor error categories lexical preference~(PR) and semantic~(SEM) show high OPUS preference rates, especially when provided with translation context (up to 62\% and 55\%, respectively), indicating that these subtler types of errors are challenging for the models. In contrast, the major error categories missing parts~(SAK) and grammatical~(GR) errors show consistently low OPUS preference rates (19--32\% and 37--45\% without context, respectively), suggesting that clear mistakes are easier to avoid.

\section{Conclusion}
\label{sec:conclusion}

We introduce the first freely available, manually annotated dataset contrasting translationese from machine translations with human-written idiomatic alternatives for Swedish. 
In creating this dataset, we conducted a detailed analysis of translationese in English-to-Swedish translations produced by both smaller specialized machine-translation systems and LLMs. 
The dataset includes the English source sentence, preceding context, a fine-grained analysis of each sample, and tags for different types of translation problems. It is a resource for studying translationese in LLM outputs and ultimately for developing models that produce more natural, idiomatic translations in non-English languages. 

Our experimental analyses reveal that the smaller multilingual LLMs we probe consistently exhibit a bias toward translationese phrasing; at best the human alternative is assigned a higher likelihood in a small majority of cases. 
Human alternatives are chosen more frequently when the English source sentence is \textit{not} provided, suggesting that models tend to follow the source closely in translations, although even without translation context, the models often prefer the translationese variant. 
Including preceding context in translation prompt guides models towards the human alternative for some samples, but the overall preference toward translationese phrasing remains strong.

\section*{Limitations}

Our dataset is based on a single source dataset and therefore lacks domain diversity. We chose subtitles as the domain because spoken dialogue is particularly challenging for machine translation, with many hard-to-translate expressions such as idioms and slang. Complementing it with other domains, particularly written ones, would however be valuable future work as it would give insights to what extent those are affected by translationese. 

As our dataset is manually constructed and includes detailed annotations of each sample, it is relatively small. While it could be expanded using methods like back-translation, this would reduce control of the samples included in the dataset. 

Choosing an appropriate set of error tags is also a trade-off. The annotators were not fully satisfied with the final tag set and sometimes found it difficult to make precise decisions. A more detailed tag set could capture finer distinctions, but it would also make annotation and interpretation more complex. 

\section*{Acknowledgments}
We thank the anonymous reviewers for their constructive feedback, which helped improve this paper. We also thank our student assistant Oskar Erkstam, who worked on the dataset annotation together with author AJ. 
This research was supported by TrustLLM funded by Horizon Europe GA 101135671. The computations were enabled by the National Academic Infrastructure for Supercomputing in Sweden (NAISS), partially funded by the Swedish Research Council through grant agreement no. 2022-06725.

\section{Bibliographical References}\label{sec:reference}

\bibliographystyle{lrec2026-natbib}
\bibliography{lrec2026-example}

\section{Language Resource References}
\label{lr:ref}
\bibliographystylelanguageresource{lrec2026-natbib}
\bibliographylanguageresource{languageresource}

\appendix

\section{A Comparison of Our Error Tags and MQM}
\label{app:mqm}

To address the similarities and differences between the MQM Error Typology and the self-developed error tags used in this article, we compare the two using the CORE Typology definitions provided on the MQM website\footnote{\url{https://themqm.org/the-mqm-full-typology/}}. Readers familiar with the MQM error classification may notice similarities between MQM categories and our tags. The purpose of this section is therefore to compare the two systems and highlight how our tagging scheme captures relationships between language type, error causes, and error effects, as well as how translation errors influence the receiver’s understanding. 

\subsection{Motivation Behind our Custom System}

Based on the MQM documentation and typology design, the framework appears primarily oriented toward organizational and technical written communication, where terminology management and adherence to conventions are central. In contrast, our error tags were designed to highlight the specific translation error patterns observed in our dataset, which consists of conversational language from the OpenSubtitles corpus.
When addressing issues in transcribed spoken dialogue, translation errors often relate to linguistic nuance, emotional tone, and idiomatic expression. These aspects play a central role in conversational language but are less emphasized in frameworks primarily designed with technical or professional translation contexts in mind. 
Although it would have been possible to extend the MQM framework with additional categories to better suit our needs, we decided that developing a custom tagging system improves the clarity of our analysis. Some of the errors we identify could be captured within the MQM framework. However, other nuances are difficult to classify without explicitly linking error causes and their effects, particularly when the analysis does not rely on MQM's scoring-based evaluation scheme. 
Our error tags were developed through an iterative annotation process. Sentences containing translation issues were manually examined, and as recurring patterns emerged, additional tags were introduced to capture these nuances. While MQM is designed to measure translation quality, our tagging system focuses on identifying systematic error patterns in the dataset. The tags allow us to capture relevant aspects of the linguistic context, identify the underlying causes of translation issues, and determine whether these issues result in loss of meaning, semantic shifts, or merely less preferable lexical choices.

\subsection{Comparison of Individual Tags}

Our \hla{\textit{Style} (ST)} tag, which captures instances of domain-specific language, is similar to the MQM \textit{Terminology} category in that both relate to the appropriate use of specialized vocabulary. Our tag covers issues corresponding to the \hlb{MQM \textit{Terminology}} subcategories (\hlb{\textit{Inconsistent with terminology resource}}, \hlb{\textit{Inconsistent use of terminology}}, and \hlb{\textit{Wrong term}}), since domain-specific communication typically requires the correct use of established terminology that can often be verified through domain resources. 
However, the role of the tag differs between the two systems.
In MQM, \hlb{\textit{Terminology}} functions as an error category and is treated as a direct cause of translation errors, typically evaluated against predefined terminology guidelines or resources. In contrast, our \hla{\textit{Style} (ST)} tag is not itself an error tag and is not tied to a specific terminology resource. Instead, it marks sentences where domain-specific language is present and where appropriate terminology or professional language conventions must be considered.
The \hla{\textit{Style}} tag therefore extends beyond terminology. It is also used to identify sentences that require attention to linguistic appropriateness and preferred forms of expression within professional or institutional contexts. 
Errors occurring in sentences marked with \hla{\textit{Style}} may therefore vary in type and severity. These may include lexically preferred ways of expressing a concept within a field (Example~\ref{ex:notguilty}), translations that only partially capture the intended meaning (Example~\ref{ex:counsel}), or cases where specialized terms are translated incorrectly, resulting in a \hla{loss of meaning (BET)}.

Our \hla{\textit{lexical error} (LF)} tag captures the same errors as the \hlb{\say{mistranslation}} subcategory of \hlb{MQM-\textit{Accuracy}}, specifically addressing cases where certain words are translated incorrectly. In cases where words are missing, aligning with the \hlb{\say{undertranslation}},\hlb{\say{omission}}, and \hlb{\say{untranslated}} subcategories of \hlb{MQM-\textit{Accuracy}}, our \hla{\textit{Missing} (SAK)} tag is used to highlight when target language sentences lack words from the source sentence or when words remain in the source language. In cases where unnecessary words are present in the translation, our \hla{\textit{Added words} (ADD)} tag is used, aligning with the \hlb{\say{addition}} subcategory of \hlb{MQM-\textit{Accuracy}}.
Our use of tags in the dataset captures causes of errors, but the outcomes can be of various types. For example, \say{Yeah, all right} may be translated simply as \say{Okej} (Okay), which is missing (\hla{SAK}) several words and changes the semantics (SEM) of the affirmative response. If the translation is strongly affected by missing, added, or incorrectly translated words, the most severe outcome is \hla{\textit{loss of meaning} (BET)}, which is tagged accordingly. If the meaning is somewhat altered and the sentence undergoes a semantic shift, the \hla{\textit{semantic} (SEM)} tag is used. If there is instead a more preferable way to express something, our \hla{\textit{Lexical preference} (PR)} tag is applied. Words may also be missing from or added to a sentence without making any substantial difference, and therefore without requiring additional tags, as shown in Example~\ref{ex:hands}. 

On a general level, the \hla{\textit{Semantic} (SEM)} tag is used to capture translations that exhibit some form of semantic shift, change in energy, or emotional nuance when compared to the source language. This may result from specific errors, but the tag can also be used independently in situations where there is no clear or severe source of the semantic shift, and the difference instead arises from the particular word choices in the translation. An example from the dataset is the response \say{Totally}, used as a positive and agreeable reply to participating in an activity, being translated as \say{Helt och hållet}. This expression is closer in meaning to \textit{completely} or \textit{entirely}, and shifts the response’s tone to something more serious than in the source language.
MQM does not provide a direct equivalent to \hla{\textit{Semantic} (SEM)} that explicitly captures semantic shifts in tone, energy, or emotional nuance, although the \hlb{\say{undertranslation}} and \hlb{\say{overtranslation}} subcategories (\hlb{MQM-\textit{Accuracy}}) partially cover shifts that have semantic effects.

Our \hla{\textit{Loss of meaning} (BET)} tag also captures \hlb{\say{mistranslation}} (\hlb{MQM-\textit{Accuracy}}), but in terms of \textit{error effect} rather than \textit{error cause}. The \hla{BET} tag is never used by itself, but instead serves as an indicator that another error is severe enough to cause the sentence to carry a different, or indistinguishable, meaning in the target language compared to the source language. 

\hla{\textit{Direct translation} (DIR)} captures literal word-by-word translations. This becomes relevant to tag because such translations can cause different types of errors, most often related to preferential or semantic issues, but they may also lead to grammatical problems or even loss of meaning in strongly idiomatic speech. No equivalent type of translational specification is found in MQM.

Our \hla{\textit{Grammar} (GR)} tag is used to capture grammatical and syntactic errors, aligning with the \hlb{\say{grammar}}, \hlb{\say{textual conventions}}, and \hlb{\say{spelling}} subcategories of \hlb{MQM-\textit{Linguistic conventions}}. Severe grammatical errors can cause \hla{\textit{Loss of meaning} (BET)} or sometimes \hla{\textit{Semantic} (SEM)} shifts if the grammatical structure affects how a sentence is interpreted. An example from the dataset is the sentence \say{You'll toast but not drink}, which is translated as \say{Du skålar, men inte dricker}, roughly meaning \textit{You toast/raise your glass, but don't drink}. The syntactic structure of the translation is incorrect and also contributes to making the phrase appear more commanding and somewhat unpleasant.

Our \hla{\textit{Idiom} (ID)} and \hla{\textit{Slang} (SL)} tags are used to identify sentences containing idiomatic or slang expressions. These are not direct errors themselves, but can be causes of different types of translation errors or issues. In MQM, the \hlb{\say{Unidiomatic style}} subcategory of \hlb{\textit{Style}} is used for situations where the target language uses grammatically correct but unnatural language. In our dataset, however, several factors related to strongly idiomatic language use are important to capture, which requires separating the type of language used from error causes and error effects.
Idioms or slang are often handled through \hla{\textit{Direct translation} (DIR)} or through \hla{\textit{lexical errors} (LF)}, and can result in \hla{\textit{Loss of meaning} (BET)}, as seen in Examples~\ref{ex:bones} and~\ref{ex:oldbone}. Other examples can also be found in the dataset, such as the slang expression \say{Nuts!}, used to express that something is crazy, which is \hla{directly translated (DIR)}. Because there is no equivalent expression in Swedish, this results in a \hla{loss of meaning (BET)}. Translating idiomatic expressions into more literal sentences may capture the general meaning, but can still cause \hla{semantic shifts (SEM)}. An example of this occurs in the dataset where the phrase \say{get well soon} is \hla{translated directly (DIR)}; in Swedish, however, this formulation becomes more assertive and commanding, which is not the intended tone when wishing someone a recovery.

Our \hla{\textit{Lexical preference} (PR)} tag is also somewhat related to the \hlb{\say{Unidiomatic style}} subcategory (\hlb{MQM-\textit{Style}}), in the sense that a lexically preferred expression may also be a more idiomatic one. However, in our dataset it is important to distinguish between cases where established idiomatic or slang expressions are used and cases where the issue concerns lexical preference from a native speaker’s perspective.

\end{document}